# TopoBERT: Plug and Play Toponym Recognition Module Harnessing Fine-tuned BERT*

Bing Zhou[a], Lei Zou[a]*, Yingjie Hu[b], Qiang Yi[c] and Daniel Goldberg[a]

[a]Department of Geography, Texas A&M University, College Station, TX;
[b]Department of Geography, University at Buffalo, Buffalo, NY;
[c]School of Geoscience, University of South Florida, Tampa, FL

## ABSTRACT

Extracting precise geographical information from textual contents is crucial in a plethora of applications. For example, during hazardous events, a robust and unbiased toponym extraction framework can provide an avenue to tie the location concerned to the topic discussed by news media posts and pinpoint humanitarian help requests or damage reports from social media. Early studies have leveraged rule-based, gazetteer-based, deep learning, and hybrid approaches to address this problem. However, the performance of existing tools is deficient in supporting operations like emergency rescue, which relies on fine-grained, accurate geographic information. The emerging pretrained language models can better capture the underlying characteristics of text information, including place names, offering a promising pathway to optimize toponym recognition to underpin practical applications. In this paper, TopoBERT, a toponym recognition module based on a one-dimensional Convolutional Neural Network (CNN1D) and Bidirectional Encoder Representation from Transformers (BERT), is proposed and fine-tuned. Three datasets (CoNLL2003-Train, Wikipedia3000, WNUT2017) are leveraged to tune the hyperparameters, discover the best training strategy, and train the model. Another two datasets (CoNLL2003-Test and Harvey2017) are used to evaluate the performance. Three distinguished classifiers, linear, multi-layer perceptron, and CNN1D, are benchmarked to determine the optimal model architecture. TopoBERT achieves state-of-the-art performance (f1-score=0.865) compared to the other five baseline models and can be applied to diverse toponym recognition tasks without additional training.

## 1 Introduction

Since the emergence of social sensing, scholars have been endeavoring to sense the pulse of society with the help of satellite images, sensor networks from IoT and various forms of textual information from the Internet. Extra attention has been paid to mining knowledge from social media because people nowadays are consciously or unconsciously sharing their views towards ongoing events online, which propels social media to become one of the few agents that reflects the real-time societal awareness, reactions and impacts of particular events. This trait is a rare feature seldom shared by other forms of data sources.

In the light of this feature, Avvenuti et al. presented an early earthquake detecting and warning system using Twitter data, which offers prompt detection of events [1]. Several case studies processed social media data with geocoding and sentiment analysis tools to analyze the spatial patterns of changing public awareness and emotions toward hurricanes in different phases of the disaster management cycle [2,3]. Huang et al. scrutinized the human mobility patterns during the COVID-19 pandemic at multiple scales based on geotagged Twitter data [4]. Zhou et al. proposed VictimFinder which is capable of harvesting social media help requests during hurricanes [5].

Let alone the fact that geographical information being one of the key elements of knowledge generation, the aforementioned studies and other similar spatial analysis and modeling are highly dependent on the location information of the social media data. However, social media users start to pay more attention to user privacy, which results in a significant drop of the number of geotagged tweets. Simultaneously, Twitter published policies forbidding users to attach precise longitudes and latitudes to tweets. Moreover, the geographical information bound up with the social media posts might not necessarily be equivalent to the place names described in the textual content of the post. Thus, extracting location information from the textual content of social media data has inevitably become an issue that needs to be addressed. This breeds the process of geoparsing, a two-step approach which includes toponym recognition (identifying place names from texts) and toponym resolution (transforming location names to geographical coordinates). This paper focuses on the first component of geoparsing.

Existing studies on toponym recognition can be categorized into four parties based on the character of the solutions, namely rule-based, gazetteer-based, statistical learning-based, and hybrid approaches. In general, statistical learning and hybrid methods that incorporate deep learning techniques render better performance than methods that solely rely on rules or gazetteers [6,7,8,9]. Based on Bidirectional Long Short-Term Memory (BiLSTM), Wang et al. introduced NeuroTPR to extract place names [6]. Qi et al. extended CoreNLP and brought about an open-sourced named entity recognition python toolkit called Stanza, which is able to detect place names and support multiple languages [7]. SAVITR is a system that combines both NLP techniques and gazetteers for real-time location extraction [8]. Hu et al. addressed the incompleteness of gazetteers and fused gazetteers, rules, and deep learning to render a reliable place name extractor, GazPNE [9].

However, those studies suffer from several limitations. First, some models do not focus only on place names, so their prediction of location name extraction might be disturbed. Second, recurrent neural network based deep learning models might suffer from information vanishing problems when the input sequence gets larger and network deeper. Third, complicated deep neural networks frequently require large, annotated datasets and are time-consuming to train to achieve promising results.

To address the aforementioned latent flaws, this paper proposes TopoBERT, a toponym recognition module based on a one-dimensional Convolutional Neural Network (CNN) and Bidirectional Encoder Representation from Transformers (BERT). It contributes in the following directions. First, several classifiers were tested and one feasible model and classifier combination based on the evaluation result of a standard dataset is determined. Second, TopoBERT was tested by an unseen dataset together with some other existing tools to verify its generalizability. Third, the tool is ready-to-use and the dataset we generated in this study can be used by other scholars to train, test, and compare different toponym recognition models and tools.

The remainder of this paper is structured as follows. The datasets involved in fine-tuning and testing the framework, a concise introduction of the holistic design of the framework, the implementation of the framework, and the parameters used in fine-tuning the framework are detailed in section 2. The results of the experiments conducted are documented in section 3. Section 4 illustrates the potential limitations of this work and lists several future research directions. Section 5 epitomizes the findings of this paper and presents the implications of this study.

## 2 Methodology

### 2.1 Datasets

Totally four different datasets were utilized to train the module and evaluate the performance. CoNLL2003 is a shared task that concerns named entity recognition, which has been widely applied to training deep learning models [10]. The data contains entities of five types: persons (PER), organizations (ORG), locations (LOC) and miscellaneous names (MISC) and other words that are irrelevant to named entities of the aforementioned four groups (O). The prefix "B-" and "I-" are used to tag the beginning of a named entity and words that fall inside a named entity [10]. The dataset is originally divided into training, validation, and test data which are noted as CoNLL2003-Train, CoNLL2003-Validation and CoNLL2003-Test. Training data is used to train a deep learning model, validation data is used to tune the hyperparameters of the model, and the test data is used to evaluate the performance of the trained model. The data distribution of each label type in the three datasets is depicted in Figures 1(a), 1(b), and 1(c), respectively. The dataset is later modified to suit the purpose of this study by labeling all the named entities as "O" except for the location entities. Around 4.1% of the tags are location entities in these datasets.

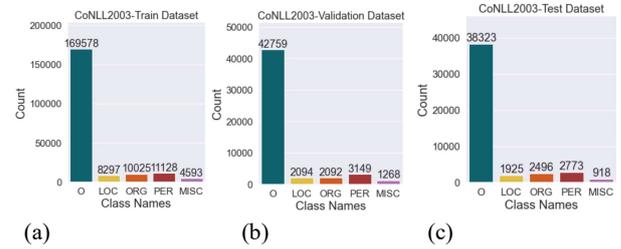

**Figure 1: Data Distribution of CoNLL2003 Dataset**

WNUT2017 is a relatively smaller dataset collected from Twitter and manually annotated, the objective of which is to tackle the issues caused by novel, emerging, singleton named entities in noisy text [11]. It aims to offer support to sustainable named entity recognition systems. This dataset contains seven different groups: person, location, corporation, product, creative work, group and none of the above. Considering the main focus of this paper and different tags used to label the dataset, this dataset is preprocessed to retain only the location entities tag and to unify the tag symbols used based on CoNLL2003 (location entities are tagged with "B-LOC" or "I-LOC" while the rest are tagged with "O"). The distribution of data under each label type in the modified dataset is shown in Figure 2(a). The total number of location names in this dataset is 1140.

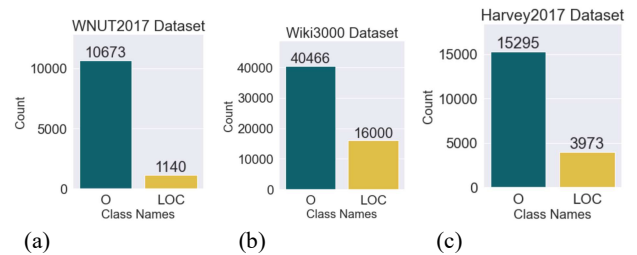

**Figure 2: Data Distribution of WNUT2017, Wiki300 and Harvey2017 Dataset**

Wiki3000 is an automatically generated dataset from Wikipedia articles by a data producing workflow proposed by Wang et al. [6]. The proposed auto-annotation approach utilizes the first paragraph of Wikipedia articles which usually encompass various entities presented with hyperlinks. These hyperlinks are later checked if they are associated with a geographical location. If so, the hyperlinked word will be labeled as a toponym. Then the Wikipedia article is divided into multiple short sentences within 280 characters with additional strategies such as random flipping to mimic the general patterns of Twitter posts [6]. The distribution of data under each label type is shown in Figure 2(b).

Harvey2017 is a dataset originally collected from the North Texas University repository (https://digital.library.unt.edu/ark:/67531/metadc993940/), which contains 7,041,866 tweets collected based on hashtag query. It was pruned, randomly subsampled and manually annotated by Wang et al. to form a new dataset with 1000 tweets aiming to evaluate NeuroTPR [6]. This dataset is adopted by this paper to test the performance of TopoBERT. The distribution of data under each label type is shown in Figure 2(c).

## 2.2 Framework Design and Implementation

As mentioned in section 1, there is an acute conflict between robust spatial analysis on social media or news media and the diminishing availability of geolocated textual context. Additionally, the location mentioned in the textual content of the tweets might differ from the geotags attached. A reliable and ready-to-use geoparser can be the mediator of such conflicts. Therefore, we present a general location extractor that can be used upon social media and news media. The workflow is shown in Figure 3.

The existing geotags of the data will be retained, and the textual contents will go through a rule-based data preprocessing module before they are fed to a zip code extractor and place name extractor. Once the place names are pulled out, a geocoding service will be applied to transform the place names into precise coordinates. The place name extractor is marked with an orange dashed rectangle in Figure 3 and serves as the crucial backbone of the entire workflow.

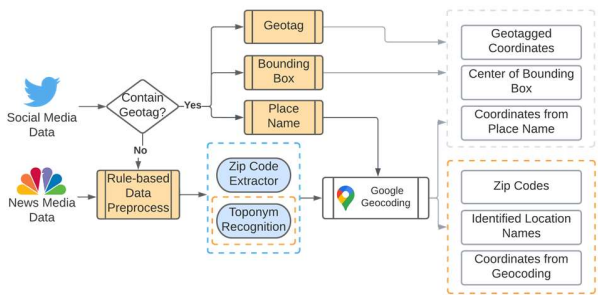

**Figure 3: Holistic Design of Location Extraction Framework for Textual Contents**

Identifying location names from input sentences is a token classification task (Figure 4), which contains two parts. A language model and a classifier. It behaves similar to how human beings analyze whether the given words are place names or not. First the language model attempts to understand the language by transforming the tokenized input data into higher dimensional space which captures the meaning of words in a given sentence, then the classifier makes predictions based on the transformed vectors and determines whether the input word belongs to location entity.

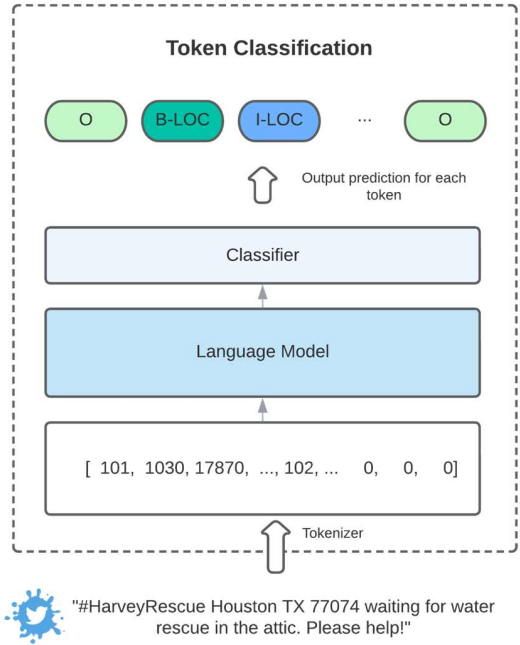

**Figure 4: Demonstration of token classification workflow.**

The heart of the proposed toponym recognition module, TopoBERT, is the Bidirectional Encoder Representation from Transformers (BERT). It is structured by stacking the encoder components of the Transformer architecture and is designed to be pretrained in an unsupervised manner. BERT takes advantage of the Attention [25] mechanism, which resolves the information vanishing issue that often upsets recurrent neural networks such as Long Short-Term Memory [26] and Gated Recurrent Neural Network [27] when the input sequence gets longer. Moreover, distinguished from many other bidirectional language models, such as ELMo designed by Peters et al. [28], in which the contextual representation of every word is the concatenation or summation of the forward and backward representations, BERT reads the entire sequence of words at once and is trained using a Masked Language Model (MLM) approach and a Next Sentence Prediction (NSP) approach which genuinely implemented the bidirectional concept or unidirectional concept. These two features combined facilitate better language understanding and bring the trophy to BERT throughout a number of NLP tasks under the General Language Understanding Evaluation (GLUE) benchmark [12].

Off-the-shelf pretrained BERT model weights can be separated into several categories based on the size of the model, whether upper and lower cases are taken into consideration, the targeted language, and unique training strategies (https://huggingface.co/transformers/v3.3.1/pretrained_models.html). Since place names are highly case sensitive and only the English language is involved in this study, 'bert-base-cased' and 'bert-large-cased' are selected as the candidate pretrained models to be evaluated. The 'bert-base-cased' model comprises 12 layers, and each hidden layer has 768 nodes, with 12 self-attention heads and a total number of 110 million parameters. The 'bert-large-cased' model consists of 24 layers, and each hidden layer has 1024 nodes, with 16 self-attention heads and 340 million parameters. The parameters are pretrained with English text from BooksCorpus (800 million words) and English Wikipedia (2,500 million words). By stacking a classifier on top of BERT, the combo can be fine-tuned to accomplish this downstream. Recent study showed that model performance can be enhanced by applying classifiers more complex than simple linear classifier or Conditional Random Field (Zhou et al. 2022). Therefore, three classifiers were examined in this study, namely linear classifier, multi-layer perceptron (MLP, Figure 5) and one-dimensional CNN (CNN1D, Figure 6). The simple linear classifier connects the output of the language model to the final prediction results with the softmax activation function. MLP applied in this study contains three fully connected layers and links the language model output with a layer with the input size equivalent to the output vector size. The number of hidden layer nodes is 256 and the output layer size equals the number of distinct labels from the training dataset. The CNN models are competent in detecting underlying features [29] and one-dimensional CNN has been successfully applied to process natural language [30, 31]. Realizing location names might share some common characteristics, the idea of CNN1D is adopted. The vector output of the language model can be considered as a one-dimensional signal and a CNN1D with kernel size 3 is applied. The output channel of the convolution is 16. Followed by a max pooling layer of size 2, which further generalizes the features and reduces model complexity. All channels of the max pooling layer output are concatenated into a single vector and is fed to a fully connected MLP with hidden layer size equals to 128.

All model combinations were implemented using Python language and pertinent packages. The dataset splitting took advantage of the ScikitLearn library and the BERT models were implemented based on the huggingface Transformer library (https://huggingface.co/transformers/). The model finetuning pipeline was built using PyTorch functions.

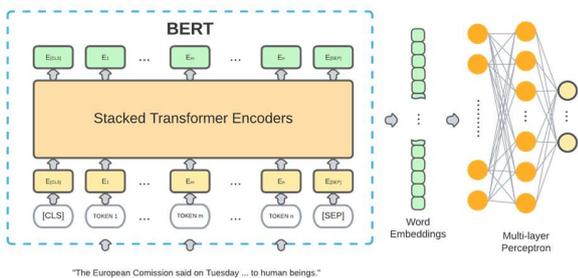

**Figure 5: TopoBERT Architecture with Multi-layer Perceptron as Classifier**

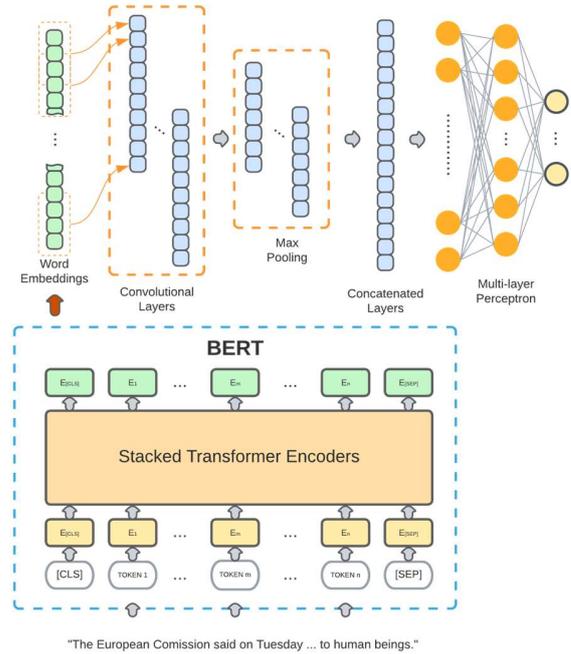

**Figure 6: TopoBERT Architecture with One-Dimensional Convolutional Neural Network as Classifier**

### 2.3 Training and Evaluation

TopoBERT is envisioned to be a ready-to-use module that renders optimal performance in toponym recognition. Models with different architectures were trained and evaluated with six datasets specified in Section 2.1 to determine the best model architecture and training strategy. The training process utilized CoNLL2003-Train as the training dataset by default and compared to another larger dataset fusing CoNLL2003, Wiki3000, and WNUT2017. The original dataset is labelled at word-level which cannot be input to BERT directly due to BERT's word-piece encoding, otherwise it will lead to large numbers of out of vocabulary words. To tackle with this issue, we first split the input data at word-level, and applied BERT word-piece tokenizer to each word. The same label was assigned to each word-piece of a single word. The labeled word-pieces are then merged to form the new input data which could be processed by BERT. This experiment aimed at measuring the performance fluctuations caused by training data size and heterogeneity. CoNLL2003-Validation was used during the training process to tune several fundamental hyperparameters such as training epochs and learning rate. CoNLL2003-Test and Harvey2017 datasets were used to evaluate the model performance. The Harvey2017 dataset was also used to benchmark TopoBERT with five prevailing toponym recognition models, namely Stanford NLP [32], spaCy (https://spacy.io/), Bidirectional LSTM-CRF [33], DM_NLP [34], and NeuroTPR [6].

The parameters of the classifier component of the module were initialized with random non-zero numbers and the BERT

component was initialized with pre-trained parameters. The entire module was trained with the fine-tuning approach [12], and the parameters were updated using a mini-batch gradient descent approach with early stopping. The maximum length of the input sequence was limited to 128 in this paper. The maximum number of training epochs was set to 50. As recommended by the original BERT paper, the initial learning rate and the training batch size were set to 2e-5 and 32 respectively [12]. Most commonly used loss function for multi-class classification task, the cross-entropy loss was employed. AdamW was selected as the optimizer during training which adjusts the learning rate dynamically to accelerate parameter convergence and implements weight decay to lower the chance of overfitting. Warm up steps, which is using a very low learning rate for the first several weight updating iterations, were also introduced during training to reduce the impact of deviating the model drastically from sudden exposure to unseen datasets.

Three commonly used evaluation metrics, precision, recall, and F1-score (Equation 1-3), were applied to gauge the performance and bias of the models. Precision calculates the percentage of correctly identified location names (noted as True Positives, TP) among all the location names predicted by the model, which combines both TP and False Positives (FP). Recall measures the percentage of correctly identified ones amongst all ground truth, which is the combination of TP and False Negatives (FN). F1-score is the harmonic mean of precision and recall, providing a comprehensive metric to evaluate model performance.

$$Precision = \frac{TP}{TP+FP}$$ (Equation 1)

$$Recall = \frac{TP}{TP+FN}$$ (Equation 2)

$$F1-score = 2 * \frac{Precision * Recall}{Precision + Recall}$$ (Equation 3)

The outputs of BERT models are at word-piece level and they are concatenated using the special prefix '##' and the word-level labels are assigned base on the starting word-piece of the word. The evaluation metrics are based on 'per-token' scores. Additionally, location name entity consists of two types of labels (B-LOC and I-LOC). In order to gauge the comprehensive performance of the model on toponym recognition, the evaluation metrics were calculated using a micro average approach, which computes a global average of precision, recall, and F1-score. It calculates the TP, FP and FN by counting the total number of TP, FP and FN under each class, namely, "B-LOC" and "I-LOC".

## 3 Results and Analysis

The first step of the experiment targeted at determining the optimal pretrained parameters for BERT model. We hypothesize that larger models outperform smaller models. To verify this hypothesis, the performance of the models initialized with 'bert-base-cased' and 'bert-large-cased' with a linear classifier stacked on top were tested. The results are displayed in Table 1.

**Table 1: Evaluation results for testing on different pretrained parameters.**

| BERT Model | Classifier | Precision | Recall | F1-score |
|---|---|---|---|---|
| bert-base-cased | Linear | 0.900 | **0.904** | 0.902 |
| bert-large-cased | Linear | **0.934** | 0.901 | **0.917** |

These two models were trained with CoNLL2003-Train and evaluated with CoNLL2003-Test. Compared to 'bert-base-cased', the precision of the prediction increased from 0.900 to 0.934 by using 'bert-large-cased' while the recall almost remained static. The F1-scores showed that 'bert-large-cased' rendered better results which is in conformity with the original BERT paper [12] and validated our initial hypothesis. Therefore, 'bert-large-cased' was harnessed in all the follow-up experiments.

The second step of the experiments aimed to measure the influence of the training data and determine the optimal classifier. The model performances were evaluated using two different datasets, CoNLL2003-Test and Harvey2017. We hypothesize that (a) the model with CNN1D classifier yield better results and (b) models trained with larger datasets perform better in placename recognition. Table 2 and Table 3 list the evaluation metrics of all the tests.

**Table 2: Evaluation results with CoNLL2003-Test dataset for testing on training data variation and classifier types.**

| Training Data | Classifier | Precision | Recall | F1-score |
|---|---|---|---|---|
| CoNLL2003 | Linear | **0.934** | 0.901 | 0.917 |
| CoNLL2003 | MLP | 0.904 | 0.910 | 0.907 |
| CoNLL2003 | CNN1D | 0.923 | **0.920** | **0.921** |
| Combined | Linear | 0.889 | 0.844 | 0.866 |
| Combined | MLP | 0.941 | 0.884 | 0.912 |
| Combined | CNN1D | **0.942** | 0.916 | 0.929 |

**Table 3: Evaluation results with Harvey2017 dataset for testing on training data variation and classifier types.**

| Training Data | Classifier | Precision | Recall | F1-score |
|---|---|---|---|---|
| CoNLL2003 | Linear | 0.895 | 0.804 | 0.847 |
| CoNLL2003 | MLP | 0.885 | 0.811 | 0.846 |
| CoNLL2003 | CNN1D | **0.898** | **0.835** | **0.865** |
| Combined | Linear | 0.872 | 0.589 | 0.703 |
| Combined | MLP | 0.932 | 0.541 | 0.685 |
| Combined | CNN1D | **0.941** | **0.668** | **0.781** |

The "CoNLL2003" under the Training Data column means CoNLL2003-Train dataset and the "Combined" represents the dataset merging CoNLL2003-Test, Wiki3000 and WNUT2017.

In Table 2, when models were trained with CoNLL2003-Train, the one with a simple linear classifier produced the best precision (0.934), and the one with CNN1D produced the best recall (0.920) and F1-score (0.921). MLP performed the worst among the three classifiers. When models were trained with a combined dataset, the model with CNN1D outperformed the rest in all three metrics with precision equal to 0.942, recall of 0.916, and F1-score of 0.929. The one with a linear classifier produced the worst results with an F1-score of 0.866. In Table 3, when models were trained with CoNLL2003-Train, the one with the CNN1D classifier outperformed the rest with precision equal to 0.898, recall of 0.835, and F1-score of 0.865. When models were trained with a combined dataset, the model with CNN1D successfully defended its trophy by rendering precision of 0.941, recall of 0.668, and F1-score of 0.781. The models with MLP worked slightly worse than the ones with linear classifiers.

The above elucidation certifies the hypothesis that models with CNN1D generate the optimal performance. It also shows that more complicated classifiers like multi-layer perceptron do not necessarily render better results.

However, when viewing Tables 2 and 3 contemporaneously, the results from training with different datasets, the metrics indicated that the model trained with the combined dataset generally performed worse than the ones trained with merely CoNLL2003-Train. This phenomenon contradicts the hypothesis that models trained with larger datasets perform better. After scrutinizing the dataset used for training, we noticed some inconsistencies in the labeling criteria of the datasets. Some examples are listed in Table 4 and the unexpected phenomenon can be interpreted by the heterogeneity of the datasets.

**Table 4: Examples of different labels across the datasets used for training the model.**

| Example Entity | Dataset | | |
|---|---|---|---|
| | CoNLL2003 | Wiki3000 | WNUT2017 |
| "Canadian" | B-MISC | O | B-LOC |
| "Planet" | O | O | B-LOC |
| "east" | O | O | B-LOC |
| "orchard" "academy" | B-ORG/ I-ORG | O | B-LOC/ I-LOC |
| "earth" | O | N/A | B-LOC |

It can be seen from Table 4 that the word "Canadian," which is labeled as "B-MISC" (beginning of a miscellaneous name), is identified as "B-LOC" (beginning of a location) in the WNUT2017 dataset. The words "Planet", "east," and "earth" are misclassified as locations in the WNUT2017 dataset. The phrase "orchard academy," regarded as an organization under the CoNLL2003 criteria, is also labeled as a location entity. In this case, combining several heterogeneous datasets can be considered adding some helpful unseen samples to the original training data while introducing a substantial amount of noise.

Rolnick et al. [13] experimented on several deep learning models when trained with noisy data and claimed that the CNN model is more resilient to noise than MLP and linear models. The trend of performance change shown in Tables 2 and 3 when trained with different datasets is in accordance with this statement. It is noticeable that the models experience an increase in precision and a drastic decrease in recall when trained with a combined dataset. This incident can as well be triggered by noisy data. Since deep learning models attempt to learn the underlying patterns of the training data, the existing noise will confuse the model, resulting in a fewer number of positive predictions. This might result in an increase in precision and a decrease in recall.

Based on the observation and interpretation above, the BERT model initialized with 'bert-large-cased', stacked with a CNN1D classifier and fine-tuned with CoNLL2003-Train was selected as the finalized TopoBERT module. Table 5 shows a comparison between TopoBERT and five other models and tools based on the Harvey2017 dataset.

**Table 5: Evaluation results with Harvey2017 dataset for comparing TopoBERT with other existing models.**

| Model | Precision | Recall | F1-score |
|---|---|---|---|
| Stanford NER (broad location) | 0.729 | 0.440 | 0.548 |
| SpaCy NER (broad location) | 0.461 | 0.304 | 0.366 |
| BiLSTM-CRF | 0.703 | 0.600 | 0.649 |
| DM_NLP | 0.729 | 0.680 | 0.703 |
| NeuroTPR | 0.787 | 0.678 | 0.728 |
| **TopoBERT** | **0.898** | **0.835** | **0.865** |

The SpaCy version v3.0 is used with model "en_core_web_sm" loaded. Broad location indicates that we include entities in both LOCATION and ORGANIZATION for Stanford NER, and we include entities in the types of LOC, ORG, FACILITY, and GPE for spaCy NER. Evaluation results show that TopoBERT prevailed in the competition with precision equals to 0.898, recall 0.835 and F1-score 0.865. This result outperformed other baseline models by at least 18%.

TopoBERT has been developed as a ready-to-use module. The output data of TopoBERT includes word labels and confidence of the prediction. It complies with JSON file format for ease of use. The source code has been uploaded to GitHub and can be accessed with the link: https://github.com/SPGBarrett/gearlab_topobert.

## 4 Discussion

This paper presents a geoparsing framework and breeds a plug and play toponym recognition module which can facilitate spatial analysis based on social media or news media data. Figure 7 shows a practical application of this framework in locating Twitter posts under fine-grained topics during hazardous events. The study area is the State of Florida, and the dots in multiple colors displayed on the map are tweets posted during Hurricane Irma harvested by

Twitter developer API. The locations of those tweets without geotags are retrieved by running TopoBERT and google geocoding service. The module also enjoys the potential of being used for location name detection for news media to pinpoint the discussed topics [14,15] and help to identify fake news [16].

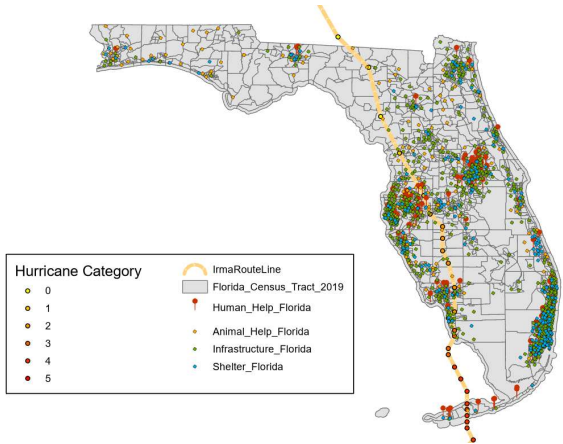

**Figure 7: Toponym recognition applied to locate Twitter posts during disasters.**

This paper concentrates mainly on designing a novel architecture of a reliable and versatile module for toponym recognition. However, the performance enhancement can continue by addressing the following issues.

First, the models are trained and evaluated based on well prepared datasets. This can be regarded as a best-case scenario compared to real life situations. Place name usage can be highly ambiguous and random, especially within social media platforms. Typos are extremely common which might cause out-of-vocabulary words in language models. Place name abbreviations such as "Boulevard" and "blvd", "Drive" and "Dr.", "Street" and "St." and so forth are frequently utilized interchangeably. People might unconsciously ignore the correct upper-case and lower-case usage, such as "college station" and "College Station", "mexico" and "MEXICO". Meticulous data preprocessing methods can be incorporated to tackle this problem in order to achieve better overall performance.

Second, several rule-base approaches can be leveraged to further boost the performance. Enlightened by the success of hybrid models [9], sets of grammar rules based on the composition of nouns, determiners, adjectives, conjunctions, numbers and possessive ending can be designed [17]. Additionally, commonly used gazetteers such as OpenStreetMap and GeoNames can be used as extra named entity matching criteria which will enhance the True Positives of the model. Regional criteria can be appended to the model while identifying place names by making country name, state names, county names, or bounding boxes as input variables of the model. This will allow the model to add constraints during processing. The top-N words from word embedding models [9,35], which are not place names, can be applied to filter words during data preprocessing. This will to some extent eliminate the False Positives of the prediction.

Third, due to the data-hungry nature of deep learning, data availability and quality are topics being inevitably discussed when large complicated deep learning models are involved. It is common knowledge in the deep learning world that larger datasets lead to better generalizability and performance. However, this statement fails to hold true in this paper due to the fact that the larger datasets are derived from several distinguished smaller datasets labeled under their own unique regime. Therefore, there is an urgent need to define criteria and build unified datasets for toponym recognition model training, evaluating and benchmarking. The dataset can be manually modified based on existing datasets and augmented using rule-based methods, gazetteers or Generative Adversarial Network [18,19,20].

Fourth, fine-tuned language models can be few-shot or zero-shot learners, which means that the models can be applied directly to certain downstream tasks with very little or even no further training [21,22,23]. This is because advanced language models can better capture the meaning of the text. This claim is also underpinned by the result of this paper which leverages BERT to boost the module capability. Therefore, incorporating gigantic models such as GPT-3 [24] might lead to another round of performance enhancement.

## 5 Conclusion

To further enhance the performance of toponym recognition by better understanding natural language, TopoBERT, which incorporate pretrained language model, BERT, is introduced. Experiments on the pretrained parameters, training dataset combinations, and model architecture reveal the following findings. First, the toponym recognition model performance is sensitive to the architecture of pre-trained language models and classifiers. The models initialized with a larger-structured BERT model ("bert-large-cased") show an advantage over the models initialized with a basic BERT model ("bert-base-cased"). More complicated classifiers like MLP do not necessarily win over simple linear classifiers. Second, increasing training data size produces worse results, especially for the recall, due to data heterogeneity. The model trained with single dataset, CoNLL2003-Train, and stacked on top with a CNN1D classifier renders the optimum results both on CoNLL2003-Test and Harvey2017 datasets. Finally, the developed TopoBERT module outperforms existing models in recognizing place names in texts. The clinched TopoBERT with the optimal model architecture and training strategy produces reliable toponym prediction and achieves F1-score of 0.865 on Harvey2017 dataset, which surpasses other prevailing models or tools by at least 18%.

In nutshell, the discoveries of this paper contribute in determining the optimal model structure on toponym recognition tasks and urges a large standardized dataset labeled with unified regime to support model training and benchmarking. A plug and play module is implemented and open sourced to support pertinent applications and similar research.

## ACKNOWLEDGMENTS

The research is supported by a project funded by the U.S. National Science Foundation: Reducing the Human Impacts of Flash Floods